\relax
\documentclass[letterpaper]{article} 
\usepackage{aaai21}  
\usepackage{times}  
\usepackage{helvet} 
\usepackage{courier}  
\usepackage[hyphens]{url}  
\usepackage{graphicx} 
\usepackage{natbib}  
\usepackage{caption} 
\usepackage[toc,page]{appendix}

\usepackage{xcolor}
\usepackage{amsmath,amssymb,amsfonts}
\usepackage{comment}
\usepackage{multirow}
\usepackage{enumitem}

\title{Spatio-Temporal Hybrid Graph Convolutional Network for Traffic Forecasting \\ in Telecommunication Networks}

\author {
        Marcus Kalander\textsuperscript{\rm 1},
        Min Zhou\textsuperscript{\rm 1},
        Chengzhi Zhang\textsuperscript{\rm 1}, 
        Hanling Yi\textsuperscript{\rm 2}, 
        Lujia Pan\textsuperscript{\rm 1}\\
}
\affiliations {
    \textsuperscript{\rm 1} Noah Ark's Lab, Huawei Technologies \ \ 
    \textsuperscript{\rm 2} The Chinese University of Hong Kong \\
    \{marcus.kalander, zhoumin27, zhangchengzhi3, panlujia\}@huawei.com, hanling.cuhk@gmail.com
}

\begin{document}

\maketitle

\begin{abstract}
Telecommunication networks play a critical role in modern society. With the arrival of 5G networks, these systems are becoming even more diversified, integrated, and intelligent.
Traffic forecasting is one of the key components in such a system, however, it is particularly challenging due to the complex spatial-temporal dependency.
In this work, we consider this problem from the aspect of a cellular network and the interactions among its base stations.
We thoroughly investigate the characteristics of cellular network traffic and shed light on the dependency complexities based on data collected from a densely populated metropolis area. Specifically, we observe that the traffic shows both dynamic and static spatial dependencies as well as diverse cyclic temporal patterns.
To address these complexities, we propose an effective deep-learning-based approach, namely, Spatio-Temporal Hybrid Graph Convolutional Network (STHGCN). It employs GRUs to model the temporal dependency, while capturing the complex spatial dependency through a hybrid-GCN from three perspectives: spatial proximity, functional similarity, and recent trend similarity. We conduct extensive experiments on real-world traffic datasets collected from telecommunication networks. Our experimental results demonstrate the superiority of the proposed model in that it consistently outperforms both classical methods and state-of-the-art deep learning models, while being more robust and stable.
\end{abstract}

\section{Introduction}
\label{sec:introduction}
Internet-connected mobile devices are penetrating every aspect of individuals’ life, work, and entertainment. 
The increasing number of smartphones and the emergence of ever more diverse applications have triggered a surge in mobile data traffic. 
As forecasted by CISCO~\cite{cisco2018cisco}, the annual worldwide IP traffic will reach 4.8 ZB by 2022 and traffic from wireless and mobile devices will account for $71\%$ of the total IP traffic by the same year. With the arrival of 5G networks, cellular networks are further moving in the direction of being more diversified, high-throughput, integrated, and intelligent. 
Thus, the networks require an intelligent supervisory system to ensure its stability and reliability. Accurate traffic forecasting is one of the key components in such as system and is vital for network planning~\cite{nie2017network}, routing configuration, and resource allocation~\cite{zhang2017traffic}. However, it is a challenging task due to 1) diverse cyclic temporal patterns within each base station and 2) complex and dynamic spatio-temporal dependencies among base stations.

A cellular network consists of a number of base stations, each covering a certain geographic area. A base station handles data traffic from users in its vicinity and relay the information to its destination. The base stations co-operate to attain a high service quality, however, their connections are not static but can be adjusted as necessary, this makes the interactions among base stations flexible, dynamic and hard to infer.

The traffic volume in a base station is dominated by the accessed users whose diverse activities and mobility bring both temporal and spatial dependencies into the cellular network traffic. The users' daily activities accounts for the temporal dependency within each individual base station~\cite{cai2016spatiotemporal}.
The spatial dependency is mainly due to user mobility, which introduces correlations into the traffic among spatially distributed base stations. As shown by~\citet{shafiq2015geospatial}, there exists spatial correlations in traffic, not only between neighboring but also distant base stations. This is partly due to the daily commute of users between residential areas and workplaces, which are typically located separately. Furthermore, the spatial dependency can change with time (i.e., it is dynamic), as revealed by our analysis in Section~\ref{sec:prelim}. 
Thus, the spatial-temporal dependency in cellular network traffic can be considerably complex and challenging to model.

Network traffic forecasting has been studied for a long time. Classical solutions model traffic patterns using either statistical time series methods~\cite{taylor2018forecasting,zhou2006traffic} or statistical learning methods~\cite{zhang2017traffic,hong2012application}. However, these solutions either consider the base stations separately or manually design features to capture the spatio-temporal correlations, and thus fail to explicitly capture the spatial dependencies and the interactions between the base stations. 
More recently, advances in deep learning enable promising solutions in modeling complex spatio-temporal correlations by 
Graph Convolutional Network (GCN) based approaches~\cite{fang2018mobile,geng2019spatiotemporal,Guo2019ASTGCN,andreoletti2019network}. 
In these works, however, the premise is a static graph structure, either known or identified to model a relevant correlation, and therefore they cannot capture any dynamic spatial dependencies. 

In this work, we investigate the characteristics of cellular network traffic and its complex and dynamic spatio-temporal correlations. This motivates the proposed STHGCN model which leverages the power of Gated Recurrent Units (GRUs) and GCNs. 

Our contributions can be summarized as follows:

\begin{itemize}
     \item We investigate the characteristics of cellular network traffic collected from an densely populated metropolitan area and shed light on its complex spatial-temporal dependencies. In particular, we observe that the cellular network traffic exhibits i) diverse cyclic temporal patterns within each base station; and ii) complex and \textit{dynamic} spatial dependencies among base stations.
    
    \item We propose a deep-learning-based model to address the observed spatio-temporal dependency. Specifically, we identify three types of spatial dependencies: spatial proximity, functional similarity and recent trend similarity (dynamic and changes with time), and design a hybrid GCN to explore the spatial information.
    Furthermore, we exploit the temporal patterns by using three different data components: recent, daily-periodic and weekly periodic. These two components are integrated seamlessly in a unified framework and trained in an end-to-end manner.
    To the best of our knowledge, this is the first work that considers both static and dynamic spatial dependencies simultaneously. 
    
    \item We carry out extensive experiments on two real-world datasets with network traffic. 
    The results demonstrate the superiority of STHGCN in telecom network traffic predictions, as it consistently outperforms both classical methods and state-of-the-art deep learning models. Benefiting from hybrid graphs, STHGCN is furthermore observed to give more stable results as compared with the baseline deep learning models.
\end{itemize}

\section{Related Work}
\label{sec:ReWork}

Cellular network traffic data is essentially spatially distributed time series data. 
Models for independent time series can have been applied~\cite{zhou2006traffic,zivot2006vector} but these are not well posed to address the correlation between the traffic of a base station and its neighbors~\cite{ghaderi2017deep,wang2018spatio}. 
Since then, spatio-temporal forecasting methods that incorporate information from neighbors have been rising in popularity~\cite{wang2015approach,zhang2017traffic}.
However, these methods are lacking in flexibility and representation power, causing researchers to look for more adaptable and powerful methods.

Inspired by success of CNNs in image processing, a line of studies have applied CNNs in their spatio-temporal models for traffic prediction tasks~\cite{ma2017learning,zhang2017deep,zhang2020citywide}.
For instance, CNNs has been used for traffic speed predictions~\cite{ma2017learning} and convolution-based residual networks has been utilized for traffic flow prediction in road networks~\cite{zhang2017deep}.
Some researchers further harness the power of CNNs and RNNs in joint deep learning models to capture the complex spatio-temporal dependency simultaneously~\cite{cheng2018deeptransport,yao2018deep,yao2019revisiting}. \citet{yao2018deep} applied a local CNN together with an LSTM model for taxi demand prediction and convLSTM and 3D CNN are combined to perform long-term mobile traffic forecasting in~\cite{zhang2018long}. 

However, CNNs and its variants are only suitable for tasks where the data has a regular structure, such as images, voice, etc. When applying CNN-based methods in traffic prediction, one common method is to preprocess the data to a grid-like structure~\cite{wang2017spatiotemporal,zhang2018long}. However, this may degrade the prediction performance due to loss of spatial granularity and can thus not be directly applied to cellular network traffic predictions since the base stations are not regularly distributed\footnote{For example, in heterogeneous cellular networks, the distance between two base stations is about 500m in an urban area, but can reach more than 2km in rural areas.}.

Recently, there has been research interest in generalizing deep learning techniques to graphs~\cite{kipf2016semi,bruna2014spectral}, which provides a good solution to tackle traffic prediction tasks where the data structure is irregular~\cite{geng2019spatiotemporal,lin2018predicting, yu2018spatio,cui2018TGCLSTM,Li2018DCRNN,zhang2019unifying,wu2019graph}. 
For instance, \citet{yu2018spatio} propose a sandwich-structured graph convolution model to use features from spatial and temporal domains, while \citet{cui2018TGCLSTM} integrate GCNs and LSTMs for roadway traffic forecasting using a graph based on the network topology. \citet{Guo2019ASTGCN} further performs highway traffic forecasting using a multi-component attention-based spatial-temporal GCN. 

Although the above methods introduced GCNs for modeling the spatial dependency, the premise is a known topology or static multi-view graphs. Since the connections in a cellular network are dynamic, these methods are not suitable to be applied directly.
In this paper, we carefully investigate the characteristics of cellular network traffic from a densely populated area and identify its unique spatio-temporal dependency, that the traffic shows both an evolving geospatial dependency and distinct daily and weekly periodicity. To properly utilize the spatio-temporal dependencies, we propose STHGCN for fine cellular network traffic forecasting. 
Specifically, the proposed model first exploits the spatial correlations between base stations using hybrid GCNs via three types of graphs: spatial proximity, functional similarity and recent trend similarity. Then GRU~\cite{chung2014empirical} is applied to capture the temporal dependency. 

\section{Preliminary Observation And Motivation}
\label{sec:prelim}

In this section, we first provide a general overview of the collected cellular network traffic dataset, and then present a detailed analysis of its temporal and spatial dependencies. This spatio-temporal dependency highlights the challenges of our traffic forecasting task and motivates our model presented in Section~\ref{sec:methodology}.

\subsection{Dataset Overview}
\label{sec:dataset}

We collected our dataset from a production LTE network with $984$ base stations in a metropolitan area of China. Most of the base stations are located in a relatively dense part of the metropolitan area of the city while a couple are located farther away.
Specifically, the dataset contains historical uplink traffic records at 15 minutes intervals with a time span ranging from $8/1/2018$ to $9/18/2018$. There are around $4.6 \times 10^6$ records in the dataset with a few examples shown in Table~\ref{tab:dataset}. 

\begin{table}[t]
\caption{Examples of traffic samples.}
\centering
\begin{tabular}{ccc}
\hline 
Time & C001 (MB) & C002 (MB)  \\
\hline 
2018-08-01 00:00:00 & 52.463 & 43.665 \\
2018-08-01 00:15:00 & 65.487 & 8.897  \\ 
2018-08-01 00:30:00 & 18.690 & 8.222  \\ 
\hline 
\end{tabular}
\label{tab:dataset}
\end{table} 

\subsection{Temporal Dependency Analysis}
\label{subsec:temporal-analysis}

\begin{figure}[t]
\begin{center}
\includegraphics[width=0.47\textwidth]{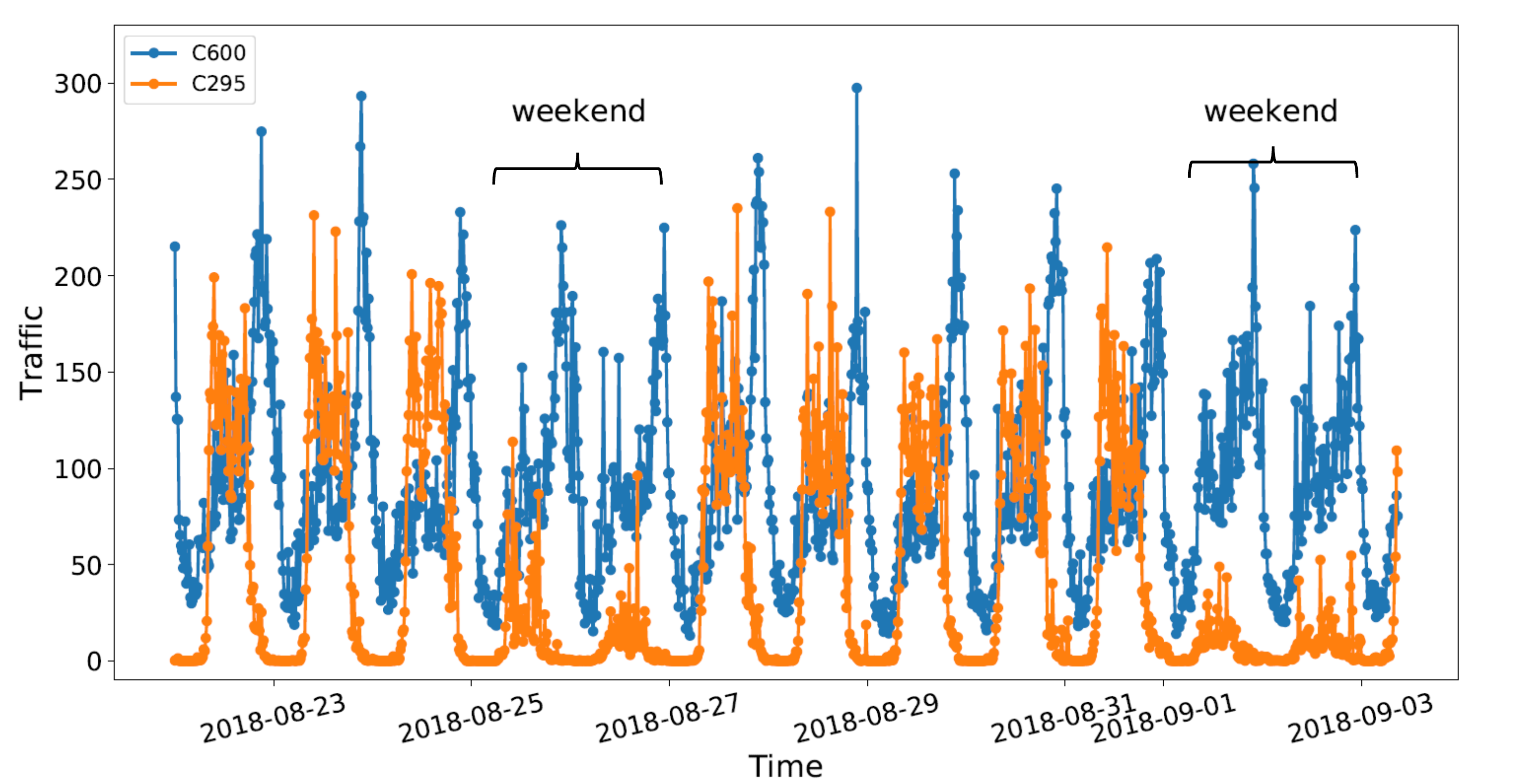}
\caption{Traffic dynamics of base stations C600 and C295.}
\label{fig:two_cells}
\end{center}
\end{figure}

To better understand the temporal characteristics of the cellular network traffic, we first investigate the temporal dynamics of the base stations. 
Two weeks of traffic data from two base stations, C600 and C295, are illustrated in Figure~\ref{fig:two_cells}.
We observe that the traffic exhibits \textit{distinct daily patterns} with low traffic volumes before dawn and heavy traffic during the daytime.
Specifically, there is a widespread intensive increase in cellular network traffic throughout the city starting from around 7:00 am, indicating that people are awake and involved in various activities.
Furthermore, we observe a peak close to midnight in most regions, which might reflect users' peculiar activity patterns in the metropolitan area where the dataset is collected from. 
We can also observe a \textit{temporal shifting of periodicity}, as the peak traffic hours in a given base station vary slightly from day to day.
In addition to the daily pattern, a \textit{strong weekly pattern} is observed in some base stations. As illustrated in Figure~\ref{fig:two_cells}, base station C295 have significantly lower traffic volumes during the weekends.

\subsection{Spatial Dependency Analysis}
\label{sec:spatial_analysis}
In this section, we introduce three types of spatial dependencies observed in our data motivating our model design.

\subsubsection{Static Dependency Analysis}
Intuitively, cellular network traffic in base stations that are close in distance are correlated due to population movements and similarity in the functions of that particular geographical area. We denote this type of spatial dependency as \textit{spatial proximity} and it is static in nature. 

In addition to the spatial proximity, it is also observed that base stations in areas that share similar functionality may have similar demand patterns. For instance, base stations located in transit centers may expect an intensive demand in the mornings and evenings during work days, and shopping malls usually have a high traffic demand during weekends. As we can observe from Figure~\ref{fig:two_cells}, the two base stations have their daily peaks at different times. This difference in daily pattern has a strong correlation with the location of the base stations. In fact, C295 is located in an industrial area, while C600 is located in a residential area. Thus, C600 has a high traffic volume during working hours while C295 has a high traffic volume during off-work hours.
We denote this type of spatial dependency as \textit{functional similarity}.
The detailed method we use to obtain the functional similarity is documented in Section~\ref{subsec:graph_construction}. Note that this is different from prior work~\cite{geng2019spatiotemporal} where the functional similarity is captured by Point of Interests (POIs) information, which is not available in this study.

\subsubsection{Dynamic Dependency Analysis}
\label{subsec:dynamic}

\begin{figure*}[t]
\begin{center}
\includegraphics[width=1.0\textwidth]{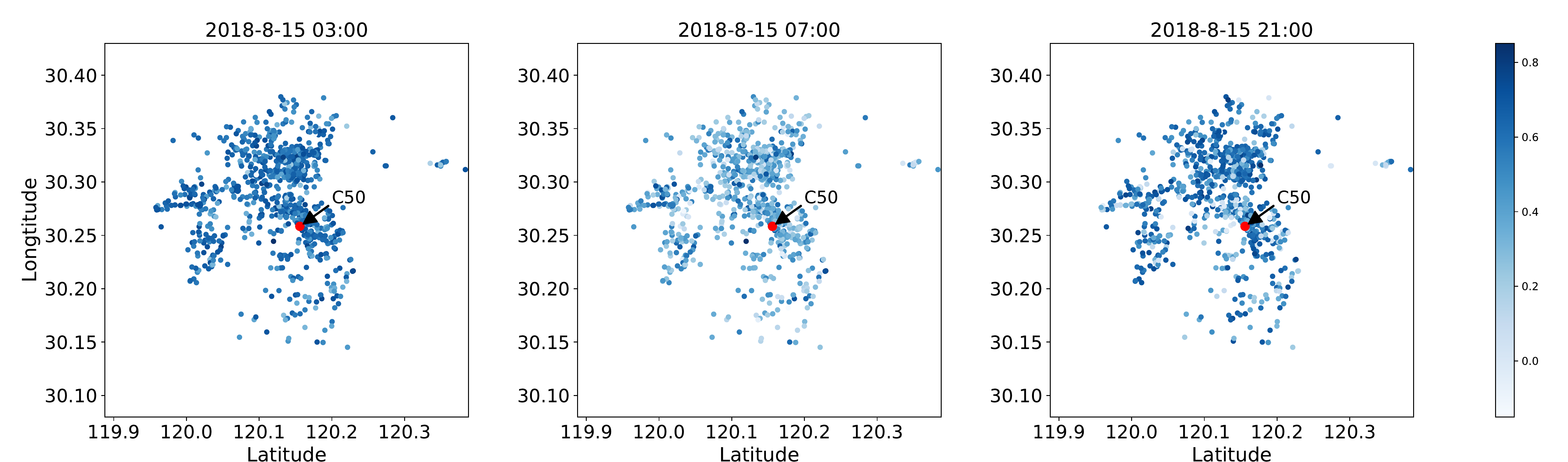}
\caption{PCC of base station C50 with all other base stations at different times of a day (best viewed in color).}
\label{fg:pcc_all}
\end{center}
\end{figure*}

Although the geographical distance and functional similarity have been considered in earlier graph-related spatio-temporal prediction methods, they only capture the static dependencies. However, dependencies between locations can change over time. This intuition motivates the definition of \textit{recent trend similarity}, which aims to capture the instantaneous correlation among base stations. To gain more insights into the dynamic correlation, we further investigate the traffic correlation of centrally located base station C50 with all other base stations at different times of a particular date. To quantify this correlation, we compute the pair-wise Pearson Correlation Coefficient (PCC) between the recent traffic\footnote{In this paper, we take the traffic in the last 12 hours as the recent traffic. Note that the recent traffic is changing with respect to time.}
volumes of two base stations, $\textbf{x}_i$ and $\textbf{x}_j$. Formally, PCC is defined as
\begin{equation}
    \rho_{\textbf{x}_i,\textbf{x}_j}=\frac{cov(\textbf{x}_i,\textbf{x}_j)}{\sigma_{\textbf{x}_i}\sigma_{\textbf{x}_j}},
\label{eq:pcc}
\end{equation}
where $cov(\textbf{x}_i,\textbf{x}_j)$ is the covariance between traffic $\textbf{x}_i$ and $\textbf{x}_j$, and $\sigma_{\textbf{x}}$ is the standard deviation of $\textbf{x}$. 

As shown in Figure~\ref{fg:pcc_all}, the pair-wise correlations of C50 with all other base stations differ significantly at different times. Thus when predicting the future network traffic, it is important to take this \textit{dynamic} geospatial dependency into account.

\section{Methodology}
\label{sec:methodology}

In this section, we start by defining the problem and related notations, we then present the details of our proposed STHGCN model. Specifically, STHGCN combines GCN and GRU to capture temporal and spatial dependencies simultaneously. We use three different data components: recent, daily-periodic and weekly-periodic data to exploit the temporal patterns. We further consider three types of spatial correlations, spatial proximity, functional similarity and recent trend similarity, and design a hybrid GCN which integrate the spatial information. An overview of STHGCN is shown in Figure~\ref{fg:overview}.

\subsection{Problem Formulation}
\label{subsec:formualtion}
In this work, the goal is to predict the uplink traffic volume of all base stations in the cellular network in a determined future period. 
Denote $\mathcal{G}^t$ as the graph set at time $t$. Each graph $G^t_i\in \mathcal{G}^t$ is built from $N$ base stations according to different criteria to capture the various spatial dependencies. We will discuss how to construct $G^t_i$ in detail later. We use $G^t_i=(V,E_i^t)$ to denote a single graph, where $V$ is the set of base stations and $E_i^t$ is the set of edges of graph $i$ at time $t$. 
Each graph $G^t_i$ is further associated with an adjacency matrix, denoted as $A_i^t\in \mathbb{R}^{N\times N}$. 
By definition, $A_i^t[j,k]$ equals to 1 if there is an edge between base station $j$ and base station $k$, and 0 otherwise. 
At time $t$, the traffic of the cellular network is denoted as $x^t\triangleq (x^{t}_1,x^{t}_2,...,x^{t}_N)\in \mathbb{R}^{N\times 1}$, where $x^{t}_i$ is the traffic volume of the $i$-th base station at time $t$.

We aim to predict the $k$-step future traffic flow at time $t$, based on the graph topology $\mathcal{G}^t$ and the historical traffic $X^t$,
\begin{equation}
   \hat{x}^{t+k}=f(\mathcal{G}^t;X^t),
\label{equ_formulation}
\end{equation}
where $X^t\in\mathbb{R}^{N\times T}$ is the historical traffic volume formally defined in Section~\ref{sec:time-slice}.

\subsection{Time Slicing to Capture Periodic Patterns}
\label{sec:time-slice}
\begin{figure}[t]
\begin{center}
    \includegraphics[width=0.45\textwidth]{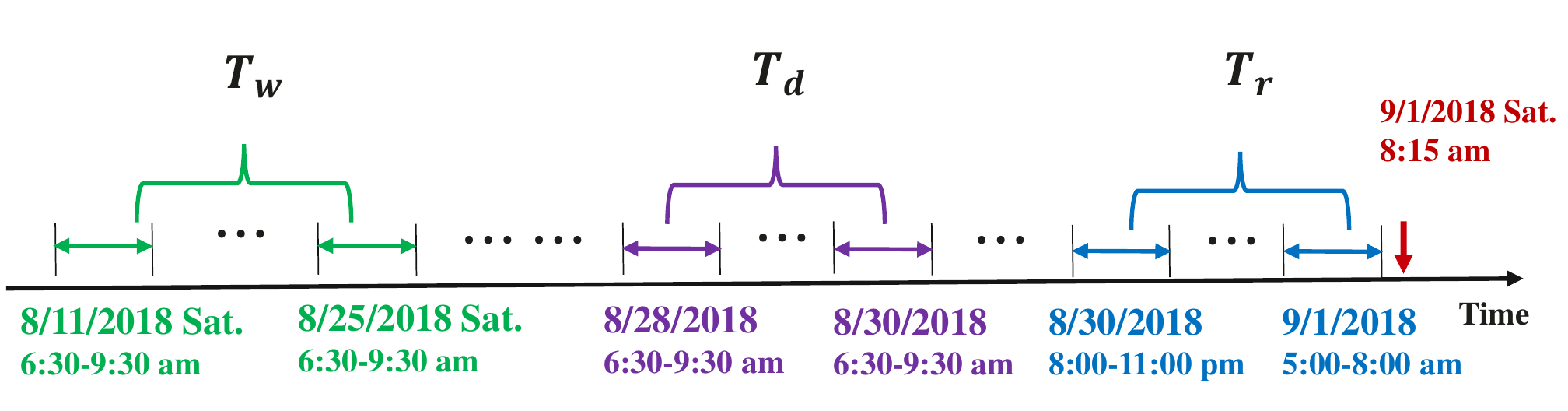}
    \caption{The time slicing using $l=12,T_r=4,T_d=3,T_w=3$ and a data granularity of 15 minutes.}
    \label{fg:timeline}
\end{center}
\end{figure}

As shown in Section~\ref{subsec:temporal-analysis}, the base stations' traffic data shows strong periodical patterns. Though recurrent neural network and its variants are promising for time series modeling, training RNNs to handle long-term information is a not easy task. The increasing length enlarges both the computational complexity and the risk of the vanishing gradient problem, thus significantly weakening the effects of periodicity~\cite{yao2019revisiting}. To address this issue, we directly utilize the periodic information and construct the inputs $X^t$ using three components consisting of the recent, daily-periodic and weekly-periodic time series segments, as illustrated in Figure~\ref{fg:timeline}. Each time series segment has a fixed length $l$ and each component can have multiple segments. The number of segments for each component are denoted as $T_r$, $T_d$ and $T_w$, respectively. Note that for the daily-periodic and weekly-periodic components, we offset the data inputs by $\lfloor l/2 \rfloor$ as illustrated in Figure~\ref{fg:timeline}, to adjust for any possible periodical shifts.
The input to the prediction problem can thus be defined as
\begin{equation}
    X^t \triangleq (X_w^t, X_d^t, X_r^t) \in \mathbb{R}^{N \times T},
\end{equation}
where $T$ denotes the historical data points we use for prediction,
\begin{equation}
    T = (T_r + T_d +T_w)*l.
\end{equation}

\subsection{Graph Construction}
\label{subsec:graph_construction}
The performance of the GCN model heavily relies on the structure of the input graph~\cite{lin2018predicting,kipf2016semi}. An adjacency matrix needs to be defined with which the graph spectral filter can be approximated.
In most previous work, the graph topology is assumed to be a premise and therefore static~\cite{geng2019spatiotemporal,zhao2019t}. However, in a cellular network, the interactions of the base stations change along with both the network configuration and user mobility.
Motivated by the observations in Section~\ref{sec:spatial_analysis}, we consider three different types of spatial correlations, including (1) spatial proximity, (2) functional similarity, and (3) recent trend similarity. Note that other types of correlations can be included to extend the model.

To determine whether two base stations should be connected in each graph, we define an indicator function $\mathcal{F}$ as,
\begin{equation}
\mathcal{F}(x) = \left\{
\begin{array}{rcl}
1,      &      & i \neq j \: \text{and} \: \left|x\right| \geq \epsilon \\
0,      &      & otherwise
\end{array} \right.
\end{equation}
where $\epsilon$ is a threshold for determining the sparsity of the adjacency matrix.
The threshold $\epsilon$ can be selected in various ways, with the most common being a fixed value. Due to the dynamic nature of part of our graphs, we instead use a relative threshold. Thus, $\epsilon$ is set to
\begin{equation}
    \epsilon = \mathcal{Q}_p(E_i^t),
\end{equation}
where $\mathcal{Q}_p$ is the quantile function and $E_i^t$ the set of weighted edges of the considered graph $G_i^t$. The threshold is thus set as to keep the number of edges fixed to a specified proportion $p$ even though the actual edge weights may vary.

\subsubsection{Spatial Proximity}
Intuitively, two base stations that are close in distance are correlated due to population movements and similarity in the functions of that particular geographical areas. In fact, spatial proximity is the most natural correlation to use, and it has been widely considered in traffic prediction tasks~\cite{fang2018mobile,zhang2019unifying}. In this work, we utilize the spherical distance between base stations to model their relationship and construct an adjacency matrix $A_{SD}$ as
\begin{equation}
A_{SD}(i,j) = \mathcal{F}\left(\exp(-\frac{d_{ij}^2}{\sigma^2})\right),
\end{equation}
where $d_{ij}$ is the geographical distance between base station $i$ and base station $j$, and $\sigma^2$ is used to control the sparsity of the matrix.

\subsubsection{Functional Similarity}
Locations sharing similar functionality may have similar demand patterns, e.g., residential areas may have a high number of demands in the morning when people transit to work, and commercial areas may expect to have high demands on weekends. Similar regions may not necessarily be close in space.
We model the functional (semantic) similarity among regions directly by their traffic similarity. To model the similarity in long term traffic demand, we consider the average weekly correlation. The adjacency matrix $A_{FS}$ is thus defined as 
\begin{equation}
A_{FS}(i,j)=\mathcal{F}\left(\rho_{x_w(i), x_w(j)}\right),
\end{equation}
where $x_w(i)$ and $x_w(j)$ are the average weekly demand pattern of the traffic in base station $i$ and $j$, respectively. $\rho_{x_w(i), x_w(j)}$ is the pair-wise correlation between $x_w(i)$ and $x_w(j)$ as defined in~\eqref{eq:pcc}.

\subsubsection{Recent Trend Similarity}
As shown in Section~\ref{subsec:dynamic}, the spatial dependency of a pair of base stations varies with time. To describe this inherent dynamic correlation, we propose to use a dynamic adjacency matrix, which captures similarities between base stations with similar short term trends. 
A straightforward way is to construct a separate graph for each timestamp based on the PCC similarity of the most recent traffic. 
We denote the recent traffic for a base station as
\begin{equation}
    \mathbf{x}^t(i) \triangleq \left(x^{t-H+1}(i),...,x^{t-1}(i),x^{t}(i)\right),
\end{equation}
where $x^{t}(i)$ is the traffic in base station $i$ at timestamp $t$, and $H$ is the number of historical data points to be considered.
The adjacency matrix $A^{t}_{RT}$ is thus defined as
\begin{equation}
    A_{RT}^{t}(i,j)=\mathcal{F}\left(\rho_{\mathbf{x}^t(i), \mathbf{x}^t(j)}\right),
\end{equation}
where $\rho_{x^t_{(i)}, x^t_{(j)}}$ is the pair-wise correlation between $x^t(i)$ and $x^t(j)$ as defined in~\eqref{eq:pcc}. Note that different from the static graphs constructed from spatial proximity and functional similarity, each input $X^t$ has its own unique \textit{dynamic} graph built with $A_{RT}^{t}$.

\subsection{Hybrid Graph Convolutional Network for Spatial Dependency Modeling}

\begin{figure*}[t]
\centering
\includegraphics[trim={3.5cm 8.5cm 10.8cm 2cm},clip,width=1.0\textwidth]{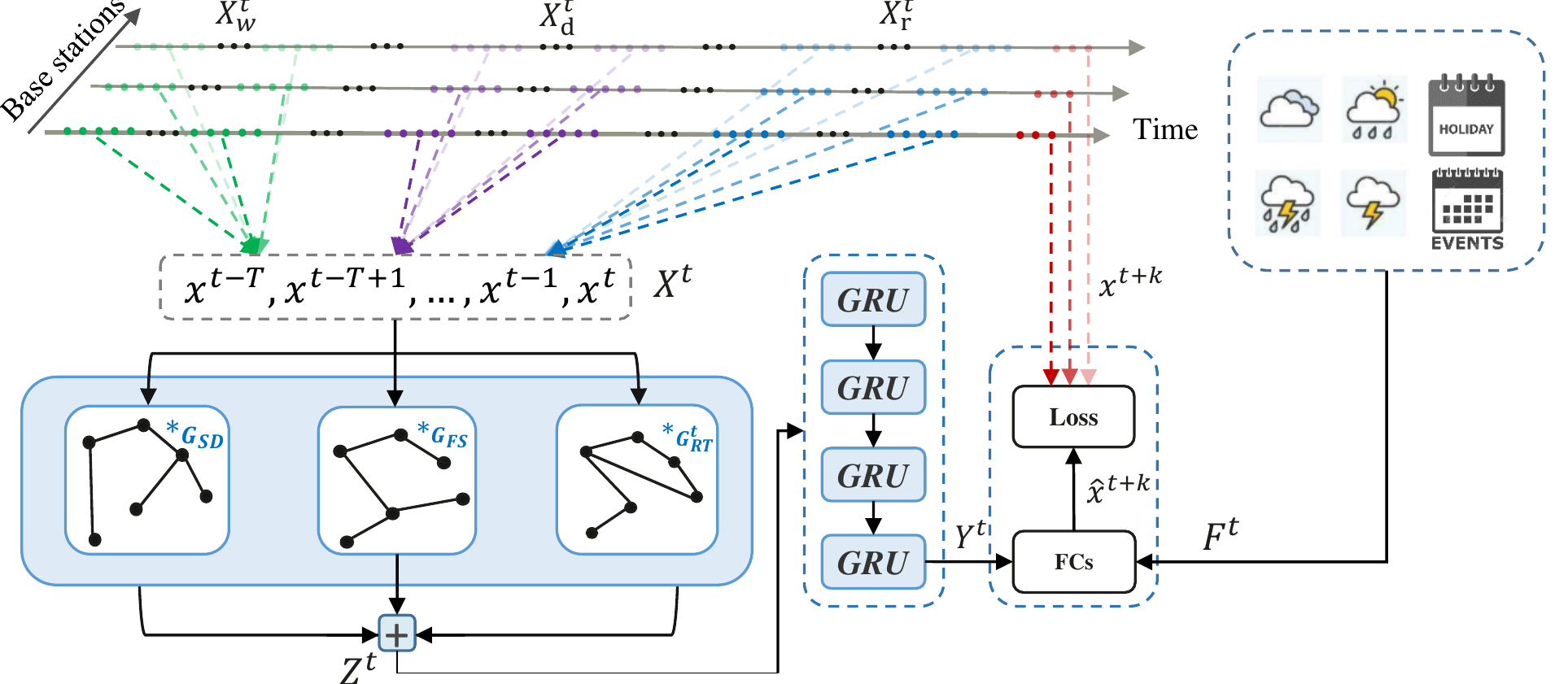}
\caption{The architecture of STHGCN. The features in each base station is sliced by time into three components to fully utilize the periodic information ($X^t$). Subsequently, $X^t$ is passed through our hybrid graph convolution where our three graphs (based on spatial proximity, functional similarity and the recent trend similarity), are applied. The result $Z^t$ is passed through a GRU, merged with any external features $F^t$, and passed through a fully connected layer to obtain the prediction $\hat{x}^{t+k}$.}
\label{fg:overview}
\end{figure*}

\subsubsection{Graph Convolutional Neural Network}
To explicitly learn the spatial dependency in cellular network traffic, we adopt the spectral graph convolutional network~\cite{bruna2014spectral} to extract the spatial features from traffic data. The idea of spectral GCN is to induce embedding features of nodes based on local structures in the Fourier domain. By doing so, we are able to consider not only the node's local behavior but also the neighborhood's general behavior.
Let $L=I-D^{-1/2}AD^{-1/2}$ denote the normalized graph Laplacian matrix, where $A$ is the adjacency matrix of a graph $G$, $D=diag(\sum\limits_{j}A_{ij})$ is the diagonal degree matrix and $I$ is the identity matrix. With the convolution operation ``$\star_{G}$'' defined in the Fourier domain on graph $G$, we calculate the convolution of a signal $X$ with a kernel $\Theta$ as
\begin{equation}
 \Theta \star_{G}X = \Theta(L)X = U\Theta(\Lambda)U^TX,
 \label{eq:graph_conv_original}
\end{equation}
where $U$ is the matrix of eigenvectors of $L$, and $\Lambda$ is the diagonal matrix of eigenvalues of $L$. 
Applying  Chebyshev polynomials approximation~\cite{defferrard2016convolutional} to localize the filter and reduce the number of parameters, we have
\begin{equation}
    \Theta(\Lambda)\approx \sum_{k=0}^K \theta_k T_k(\tilde{\Lambda}),
\end{equation}
where $\tilde{\Lambda}\triangleq 2\Lambda/\lambda_{max}-I$ is a rescaled version of $\Lambda$, $\lambda_{max}$ denotes the largest eigenvalue of $L$, $\theta\in\mathbb{R}^K$ is a learnable vector of polynomial coefficients, and $K$ is the kernel size of the graph convolution that determines the maximum radius of the convolution from central nodes. 
With these approximations, we rewrite the graph convolution in \eqref{eq:graph_conv_original} as
\begin{equation}
 \Theta \star_{G}X=\Theta(L)X=\sum_{k=0}^{K-1}\theta_k T_k(\tilde{L})X,
 \label{eq:graph_conv}
\end{equation} 
where $\tilde{L}\triangleq2L/\lambda_{max}-I$.

\subsubsection{\textbf{Hybrid Graph Convolution}}
With the constructed static and dynamic graphs, we propose a hybrid graph convolution network to integrate their spatial information.
At time $t$, we have a graph set with $m$ graphs, denoted by $\mathcal{G}^t=\{G_1^t, G_2^t, \cdots, G_m^t\}$. Note that here the graphs can be dynamic, i.e., we allow the adjacency matrix of the graphs to change with time.
We then implement a one-layer ChebyNet on each graph, specifically, for graph $G_i^t$, we obtain the convolution features as
\begin{equation}
    Z_i^t=ReLU(\Theta \star_{G_i^t}X^t)\in\mathbb{R}^{N\times c_h},
\end{equation}
where $\star_{G_i^t}$ is the convolution operator defined on graph $G_i^t$,
$X^{t}\in\mathbb{R}^{N\times T}$ is the input features, and $c_h$ is the dimension of the output.
The final output of hybrid-graph ChebyNet, denoted by $Z^t$, is obtained by combining outputs of the different graphs together as
\begin{equation}
    Z^t = \sum_{i=1}^m Z_i^t \in\mathbb{R}^{N\times c_h}.
\end{equation}
Consequently, we propagate information between base stations according to multiple graphs, and achieve node representations in the spatial domain. 
We note that by definition the hybrid GCN is similar to multi-graph convolution operator defined in \cite{geng2019spatiotemporal}. However, their considered graphs are static. Here we have both static and dynamic graphs, and thus use ``hybrid'' to emphasize this difference.

\subsection{Gated Recurrent Unit for Temporal Dependency Modeling}
Traffic data has a dynamic temporal dependency due to human behavior and user mobility. To model the temporal correlation, we utilize Gated Recurrent Units (GRUs)~\cite{chung2014empirical}. 
As a variant of RNN, a GRU uses gating information to prevent the vanishing gradient problem. 
Benefiting from its parsimonious structure, the performance of a GRU is on par with LSTM while being more computationally efficient. In our model, the parameters of the GRU are shared across all base stations to reduce model complexity. Given the sequential input $Z^t\in\mathbb{R}^{N\times c_h}$, we take the last hidden state of the GRU as its output and denote it as $Y^t\in \mathbb{R}^{N\times C}$.

\subsection{Our Model: STHGCN}
In summary, in STHGCN the input $X^t$ is passed through hybrid-graph ChebyNet to obtain $Z^t$ that capture the spatial dependency, and consecutively through GRUs to obtain $Y^t$ to capture the temporal dependency. 
$Y^t$ is concatenated with any external features $F^t$. These can be information regarding weekday, weather, etc.
We obtain the $k$-step prediction by passing the concatenated $Y^t$ and $F^t$ through a fully connected layer,
\begin{equation}
    \hat{x}^{t+k} = [Y^t, F^t]W_f + b_f,
\end{equation} 
where $W_f\in \mathbb{R}^{(C+|F|)\times 1}$ and $b_f\in \mathbb{R}^{N \times 1}$ are learnable parameters.
An overview of STHGCN is shown in Figure~\ref{fg:overview}. 

The model is trained by minimizing the combination of Mean Square Error (MSE) and Mean Absolute Error (MAE). Namely, the loss function is defined as 
\begin{align}
    L(\hat{x}^{t+k},x^{t+k}) = & \sum_{t\in T_{train}} \frac{1}{N}\sum_{i=1}^N \left(\hat{x}^{t+k}_i-x^{t+k}_i\right)^2 \nonumber \\ & + \alpha\frac{1}{N}\sum_{i=1}^N |\hat{x}^{t+k}_i-x^{t+k}_i|,
\label{eq:loss_fun}
\end{align}
where $\alpha$ is the input parameter that balances the importance between MSE and MAE.
\section{Experiments}
\label{sec:evaluation}
In this section, we validate the proposed STHGCN model with extensive experiments to demonstrate the prediction performance of STHGCN against six baselines. We begin with a short introduction to the datasets and the preprocessing steps.

\subsection{Dataset Description}
\textbf{CelluarHZ} denotes our cellular network traffic dataset, for a detailed description readers can refer to Section~\ref{sec:dataset}. The data from $8/1/2018$ to $9/4/2018$ is used as the training set, $9/5/2018$-$9/11/2018$ as the validation set and $9/12/2018$-$9/18/2018$ as the testing set.

\textbf{Abilene} is a publicly accessible link traffic dataset gathered from a backbone network in America~\cite{Zhang2004Abilene}. It contains 12 nodes and 15 bidirectional links.
Similar to~\citet{andreoletti2019network}, we treat the links as nodes and aim to predict the traffic load on the links in both directions.
In this way, we obtain a graph with 30 nodes.
Since the physical network topology of the Abilene network is known, we substitute the spatial proximity graph with the true topology, however, the functional similarity and recent trend similarity graphs are not changed.
The data from $5/1/2004$ to $7/29/2004$ is selected as the training set, and $7/30/2004$ to $8/19/2004$, $8/21/2004$ to $9/10/2004$ as the validation set and testing set, respectively. The data is downsampled to a time granularity of 15 minutes.
Different from a cellular network, each node of a backbone network aggregate more traffic volume and is more stable. Together with the provided topology this makes for a slightly easier prediction problem.

\subsection{Experimental Settings}
\subsubsection{Preprocessing}
Missing values are filled with the average value from the same time in previous periods. The data is normalized using the z-score method based on the training data. During evaluation, the predicted values are re-scaled back to the normal values.

\subsubsection{Baselines}
We compare our STHGCN model with the following six baselines, covering both classical and deep learning based algorithms. We further include an STHGCN version without additional external features.

\begin{itemize}[itemsep=0.5pt]
    \item \textbf{Historical Average (HA)}~\cite{liu2004summary} predicts the traffic data of each base station by using the average of the historical traffic in the same corresponding time slots in previous periods.
    \item \textbf{Prophet}~\cite{taylor2018forecasting} predictions using an additive model based on time series decomposition.
    \item \textbf{Spatio-Temporal GCN (STGCN)}~\cite{yu2018spatio} models both spatial and temporal patterns using graph convolution networks.
    \item \textbf{Attention Based Spatial-Temporal GCN (ASTGCN)}~\cite{Guo2019ASTGCN} forecasts traffic using a multi-component attention-based spatial-temporal graph convolutional network.
    \item \textbf{Diffusion Convolutional Recurrent Neural Network\\ (DCRNN)}~\cite{Li2018DCRNN} integrates diffusion convolution operation and Seq2Seq architecture for spatial-temporal forecasting.
    \item \textbf{Graph WaveNet}~\cite{wu2019graph} integrates diffusion graph convolutions with 1-D dilated convolutions.
    \item \textbf{STHGCN-NH} our model but without the auxiliary external feature module. Added for fairness to the other baselines that do not have the additional input.
 \end{itemize}

In our experiments, we are interested in the prediction task for $[1,2,3,4]$-steps ahead. Namely, for any given time, we want to predict the value in the next $h$ minutes, where $h\in [15,30,45,60]$.
All the deep learning baseline models use 48 samples as input and are trained to minimize the one-step prediction error. For multi-step predictions, we use the predicted values in previous steps as part of the input. 
Note that directly minimizing the loss for steps $k>1$ could potentially give better results.


\subsubsection{Hyperparameters} We train our model by minimizing the loss function given by~\eqref{eq:loss_fun} with $\alpha=10^{-4}$ using the RMSProp optimizer for $50$ epochs.
The learning rate is initially set as $10^{-3}$ and it decays exponentially every $5$ epochs with decay rate $0.7$.
We set $l=12$, $T_r=T_w=1$ and $T_d=2$. For the GRU, we set its hidden dimension to be $C=32$, and for ChebyNet, we set $K=3$ and $c_h=64$.
The adjacency matrix threshold $\epsilon$ is set to keep the edges with weights above the 90th percentile.
$\sigma$ for the spatial proximity graph is set to be $100$, and for the recent trend we set $H$ equal to $48$.
We use information regarding holidays and weekends as external features $F^t$.

\subsubsection{Evaluation Metrics}
We evaluate the prediction performance with two metrics, Root Mean Square Error (RMSE) and Mean Absolute Error (MAE), defined as follows
\begin{align}
    RMSE &= \sqrt{\frac{1}{N_T N}\sum_{t=1}^{N_T}\sum_{i=1}^{N}\left(\hat{x}^t(i)-x^{t}(i) \right)^2},\\
    MAE &= \frac{1}{N_T N}\sum_{t=1}^{N_T}\sum_{i=1}^{N}\left|\hat{x}^{t}(i)-x^{t}(i) \right|,
\end{align}
where $\hat{x}^{t}_i$ and $x^{t}_i$ are the predicted and real value, respectively, of $i$-th base station at time $t$. $N$ is total number of base stations and $N_T$ is the number of time steps.

\subsection{Experimental Results}
\subsubsection{Comparison with Baselines} 

\begin{table*}[t]
\centering
\caption{Test accuracy (average over 10 runs) on CellularHZ and Abilene.}
\label{tab:results}
\small
\begin{tabular}{cc|cc|cc|cc}
\hline
\multicolumn{1}{c}{\multirow{2}{*}{Dataset}} &
\multicolumn{1}{c|}{\multirow{2}{*}{Model}} & \multicolumn{2}{c}{15 min} & \multicolumn{2}{|c}{30 min} & 
\multicolumn{2}{|c}{60 min} \\
\cline{3-8}
& 
& \multicolumn{1}{c}{RMSE} & \multicolumn{1}{c}{MAE} & \multicolumn{1}{|c}{RMSE} & \multicolumn{1}{c}{MAE} & 
\multicolumn{1}{|c}{RMSE} & \multicolumn{1}{c}{MAE} \\
\hline
                   
\multirow{6}{*}{CelluarHZ}
  & HA & 108.9 & 44.2 & 
  108.9 & 44.2 &  
  108.9 & 44.2 \\

  & Prophet &  70.3 & 42.3 & 
  72.2 & 43.0 & 
  75.2 & 44.2 \\

  & STGCN & $52.7\pm1.0$ &$24.7\pm 0.2$  &
  $66.8\pm1.6$&$30.8\pm0.4$&
  $82.4\pm2.5$&$37.9\pm0.8$\\

  & ASTGCN & $72.4 \pm 2.7$ & $28.5 \pm 1.2$  & $74.5 \pm 2.7$ & $29.9 \pm 1.2$ & 
  $82.1 \pm 2.7$ & $35.2 \pm 1.3$ \\
  
  & DCRNN & $52.6\pm0.3$ &$24.4\pm 0.1$ &
  $64.1\pm0.7$&$29.8\pm0.2$&
  $78.5\pm1.2$&$36.5\pm0.5$ \\
   
  & Graph WaveNet & $51.7 \pm 0.2$ & $22.7 \pm 0.2$  &
  $59.0 \pm 1.12$ & $27.1 \pm 0.4$  &
   $73.7 \pm 2.0$&  $32.8 \pm 0.9$  \\
  
  & STHGCN-NH 
  &$\mathbf{47.1\pm0.2}$ &$\mathbf{21.7\pm 0.1}$ 
  & $\mathbf{54.3\pm0.3}$&$\mathbf{25.5\pm0.1}$
  & $64.9\pm0.5$ & $30.7\pm0.2$ \\
  
   & \textbf{STHGCN} & 
   $47.2\pm0.1$ &$\mathbf{21.7\pm 0.1}$ &
   $\mathbf{54.3\pm0.2}$&$\mathbf{25.5\pm0.1}$&
   $\mathbf{64.2\pm0.4}$&$\mathbf{30.4\pm0.2}$ \\
  
  \hline
 \multirow{6}{*}{Abilene}
  & HA & 53.4  & 30.1   & 
  53.4  & 30.1   &
  53.4  & 30.1   \\

  & Prophet & 31.1  & 17.8   & 
  32.3  & 18.5   & 
  34.2  &  19.7   \\

  & STGCN & $23.7\pm0.2$ &$12.9\pm 0.1$ &
  $28.9\pm0.3$&$15.8\pm0.2$&
  $34.2\pm0.6$&$19.0\pm0.3$\\

  & ASTGCN & $24.0 \pm 0.3$ & $12.9 \pm 0.1$  &
  $29.9 \pm 0.4$ & $16.2 \pm 0.1$  &
  $34.7 \pm 0.4$ & $19.3 \pm 0.2$  \\
  
  & DCRNN & $22.4\pm0.2$ &$11.9\pm 0.04$  &
  $27.6\pm0.2$&$14.5\pm0.09$&
  $32.7\pm0.3$&$17.1\pm0.1$ \\

  & Graph WaveNet & $\mathbf{22.1 \pm 0.3}$ & $\mathbf{11.7 \pm 0.13}$  &
  $26.9 \pm 0.4$ & $14.1 \pm 0.11$  &
  $31.5 \pm 0.4$ & $16.4 \pm 0.15$  \\
  
  & STHGCN-NH & 
  $22.3 \pm 0.3$ & $11.8 \pm 0.10$ &
  $26.5 \pm 0.2$ & $14.1 \pm 0.05$ &
  $30.3 \pm 0.3$ & $\mathbf{16.1 \pm 0.08}$ \\ 
  
  & \textbf{STHGCN} & 
  $22.2 \pm 0.3$ & $11.8 \pm 0.10$ &
  $\mathbf{26.3 \pm 0.2}$ & $\mathbf{14.0 \pm 0.08}$ &
  $\mathbf{30.2 \pm 0.4}$ & $\mathbf{16.1 \pm 0.18}$ \\
  \hline
\end{tabular}
\end{table*}

For the deep learning models, we run the experiments for 10 times and report their mean and standard deviation. 
Table~\ref{tab:results} summarizes the prediction results on the two datasets. As shown, STHGCN achieves almost always the best performance on both datasets for all values of $k$, independent of the metric used.
In particular, the results of the univariate time series analysis methods are not ideal, which can be explained by their inability to capture the spatial dependency. This demonstrates the limitation of univariate time series analysis methods in modeling spatio-temporal data, and emphasizes the importance of modeling spatial dependency. 
Furthermore, STHGCN outperforms the state-of-the-art deep learning baselines with a considerable margin. Specifically, compared to the second base deep learning model, STHGCN achieves an 15\% higher RMSE and 7\% higher MAE for the 4-step ahead prediction task on our CellularHZ dataset.
Different from CellularHZ, Abilene exhibits more stationary traffic flow on each node and the dependencies are more stable.
STHGCN still obtains good results, demonstrating the effectiveness and versatility of STHGCN in traffic predictions on telecommunication networks.

Compared with other deep learning methods, we also observe smaller standard deviations for STHGCN indicating more consistent predictions. This further demonstrates the robustness and stability of STHGCN, which are critical considerations in real-world systems.

The addition of the external features into the model gives minor performance gains for larger $k$ on the CellularHZ dataset. For Abilene, there is even less benefit, which may be due to it being more stable.
Our datasets relatively short time range could contribute to diminishing the effectiveness of the external feature module. However, we can conclude that the additional input does not give STHGCN an unfair advantage over the other baselines.

\begin{table}[t]
\centering
\caption{Performance comparison of STHGCN variants on CellularHZ (4-step ahead prediction).}
\begin{tabular}{c|c|c}
  \hline 
  Model Variants & RMSE & MAE \\
  \hline
  No time slicing ($T_w=T_d=0$) & $69.6$ & $33.0$  \\
  Spatial proximity only & $65.6$ & $32.9$ \\
  No dynamic traffic similarity & $64.3$ & $30.7$  \\
  STHGCN & $\mathbf{63.5}$ & $\mathbf{30.2}$ \\
  \hline
\end{tabular}
\label{tab:model_compare}
\end{table}

\subsubsection{Benefits of Time Slicing and Hybrid Graphs}
To further investigate the effect of time slicing and the advantage of hybrid graphs, we evaluate different variants of STHGCN by removing either time slicing (i.e., setting $T_w=T_d=0$ while increasing $T_r$ to keep the size of the input the same) or different types of graphs from the model. We run the experiments for 10 times and report the results with the \textit{best} 4-step ahead prediction in Table~\ref{tab:model_compare}.

We make the following observations: (1) Removing time slicing of the input significantly degrades the model performance. This highlights the benefit of incorporating historical data that capture seasonal patterns. (2) Removing any of the graph components causes performance degradation which highlights the importance of each correlation type.
In particular, by adding functional similarity and recent trend similarity, we can observe 3\% improvement in RMSE and 9\% in MAE when predicting 4-step ahead traffic. All graphs thus encode important knowledge which is leverage for more accurate predictions. 
\section{Conclusion}
\label{sec:conclusion}
In this paper, we investigate the cellular network traffic forecasting problem and identified its unique spatio-temporal dependency.
We propose STHGCN, a deep learning model combining GRU and GCN to simultaneously capture the temporal and spatial dependencies. The model takes advantage of the temporal dependency through different components (recent, daily-periodic and weekly-periodic), and simultaneously exploits the spatial dependency through three separate angles: spatial proximity, functional similarity and recent trend similarity. 
When evaluated on two real-world telecommunication traffic datasets, the proposed approach achieved significantly better results than state-of-the-art baselines. It is worth noting that the proposed STHGCN is motivated by the traffic dynamics of cellular networks, and is therefore highly suitable for traffic prediction on 5G LTE networks.
For future work, we plan to investigate the following aspects: (1) apply STHGCN to other datasets (e.g., road traffic) to verify its versatility (2) investigate more in-depth on how to best model the dynamic spatio-temporal dependency, (3) study the traffic prediction with evolving networks.
\bibliography{paper}

\end{document}